%% file: ijcnlp2017.tex
\documentclass[11pt,letterpaper]{article}
\usepackage{ijcnlp2017}
\usepackage{times}
\usepackage{latexsym}
\usepackage{graphicx}
\usepackage{url}
\usepackage{amsmath}
\usepackage{tikz}
\usepackage{pgfplots}
\usepackage{amsmath}
\usepackage{array,ragged2e}

\ijcnlpfinalcopy

\title{Boosting Neural Machine Translation}

\author{
  Dakun Zhang \ \ {\normalfont and} \ \ Jungi Kim \ \  {\normalfont and} \ \  Josep Crego  \ \ {\normalfont and} \ \ Jean Senellart \\
  {\tt firstname.lastname@systrangroup.com} \\
  SYSTRAN / 5 rue Feydeau, 75002 Paris, France \\
}

\date{}

\begin{document}

\maketitle

\begin{abstract}
Training efficiency is one of the main problems for Neural Machine Translation (NMT).
Deep networks need for very large data as well as many training iterations to achieve state-of-the-art performance.
This results in very high computation cost, slowing down research and industrialisation.
In this paper, we propose to alleviate this problem with several training methods based on data boosting and bootstrap with no modifications to the neural network.
It imitates the learning process of humans, which typically spend more time when learning ``difficult'' concepts than easier ones.
We experiment on an English-French translation task showing accuracy improvements of up to $1.63$ BLEU while saving $20\%$ of training time.
\end{abstract}

\section{Introduction}

With the rapid development of research on Neural Machine Translation (NMT), translation quality has been improved significantly compared with traditional statistical based method \cite{bahdanau2014neural,cho2015using,zhou2016deep,sennrich2015neural}.
However, training efficiency is one of the main challenges for both academia and industry.
A huge amount of training data is still necessary to make the translation acceptable \cite{koehn2017six}.
Though new techniques have recently been proposed \cite{1706.03762,gehring2017convolutional},
fully trained NMT models still need for long training periods (sometimes by weeks) even using cutting-edge hardware.

NMT system directly models the mappings between source sentence $x_{1}^{n}=(x_{1},...,x_{n})$ and target sentence $y_{1}^{m}=(y_{1},...,y_{m})$, with $n$ and $m$ words respectively \cite{sutskever2014sequence}.
Usually, such system is  based on an encoder-decoder-attention framework, in which the source sentence is fed into an encoder word by word to form a fixed length representation vector, with a forward sequence of hidden states $(\overrightarrow{h_{1}},...,\overrightarrow{h_{n}})$ and a backward sequence $(\overleftarrow{h_{1}},...,\overleftarrow{h_{n}})$.
With the attention mechanism \cite{bahdanau2014neural}, a decoder is used to decide which part of the source sentence to pay attention to and predict corresponding word representation at time $t$ together with history predictions before time $t$.
Then, a softmax is used to restore the word representation to natural target words.
During the training process, the parameter $\Theta$ is optimized:
\begin{center}
\(
p(y_{1}^{m}|x_{1}^{n};\Theta) = \prod_{t=1}^{m} p(y_{t}|y_{<t},x_{1}^{n};\Theta)
\)
\end{center}

The amount of parameters which is proportional to the network size and the size of training corpora both decide the cost of training for NMT systems. 
In order to achieve an acceptable performance on systems, deep networks (up to 8 layers) and more iterations (10-18 epochs) are necessary with a certain amount of data \cite{wu2016google,DBLP:journals/corr/CregoKKRYSABCDE16}.
Since several weeks are needed to generate results, it is difficult to experiment with several different meta-parameters, hence slowing down innovation.

While \newcite{wu2016google} proposes a brute-force approach with massive data and model parallelism as a way to accelerate training, in this paper,
we focus on a different approach based on ranking training sentence pairs by ``difficulty''.
We aim at boosting the optimisation problem through targeting difficult training instances rather than spending time on easier ones.

Every several epochs, we re-select 80\% of the data from the corpus with the highest perplexity (ppl.) to use for training.
There is no extra calculation cost since we get these ppl. loss from the previous epoch.
Finally, we achieve a 1.63 BLEU points improvement while saving 20\% training cost at the same time, which is quite a stable improvement compared with our baseline system on English-French translations.

\section{Related Work}

Training efficiency has been a main concern by many researchers in the field of NMT.
Data parallelism and model parallelism are two commonly used techniques to improve the speed of training \cite{wu2016google}.
As a result, multiple GPUs or TPUs\footnote{Google's Tensor Processing Unit \cite{wu2016google}.} are needed which requires additional replication and combination costs.
Data parallelism does not reduce the actual cost of computation, it only saves time by using additional computational power.

\newcite{chen2016efficient} proposed a modified RNN structure with a full output layer to facilitate the training when using a large amount of data.
\newcite{kalchbrenner2016neural} proposed ByteNet with two networks to encode the source and decode the target at the same time.
\newcite{gehring2017convolutional} use convolutional neural networks to build the system with a nature of parallelization.
\newcite{kuchaiev2017factorization} focus on how to reduce the computational cost through patitioning or factorizing LSTM matrix.
To compare, our method does not modify the network itself and can be used in any NMT framework.

Other methods focus on how to reduce the parameters trained by the model \cite{see2016compression}. They show that with a pruning technique, 40-60\% of the parameters can be pruned out.
Similar methods are proposed to reduce the hidden units with knowledge distillation \cite{DBLP:journals/corr/CregoKKRYSABCDE16,kim2016sequence}.
They re-train a smaller student model using text translated from teacher models.
They report a 70\% reduction on the number of parameters and a 30\% increase in decoding speed.
\newcite{hubara2016quantized} proposed to reduce the precision of model parameters during training and network activations, which can also bring benefits to training efficiency.

To the best of our knowledge the closest idea to our work is instance weighting \cite{jiang2007instance}, which is often used for domain adaptation.
They add instance dependent weights to the loss function to help improving the performance.
As a comparison, we focus on using ``difficult'' instances in training rather than spending training time on easier ones.
We improve the accuracy while simultaneously reducing the cost of training.

\section{Training Policies}

To train a NMT system, we first design a neural network with some fixed meta-parameters (e.g. number of neurons, LSTM layers, etc.).
Then, we feed the network with training instances (in a way of mini-batch) \cite{pmlr-v37-ioffe15} until all instances of a training set are consumed.
The operation is repeated until the model converges.
Parameters of the network turn to an optimum status and as a consequence, the system reaches best performance.

During this process, there is no distinction between training instances.
That is, all the instances are regarded as equal and used to train the NMT model equally.
However, some instances are relatively easy for the model to learn (e.g. shorter or frequently used sentences). To use them repeatedly results in a waste of time.
Moreover, to avoid overfitting and to help convergence, training instances are often randomly shuffled before used to train a model,
hence, introducing un-certainty to the final status of the system.

\begin{figure}[h!]
   \includegraphics[width=0.48\textwidth]{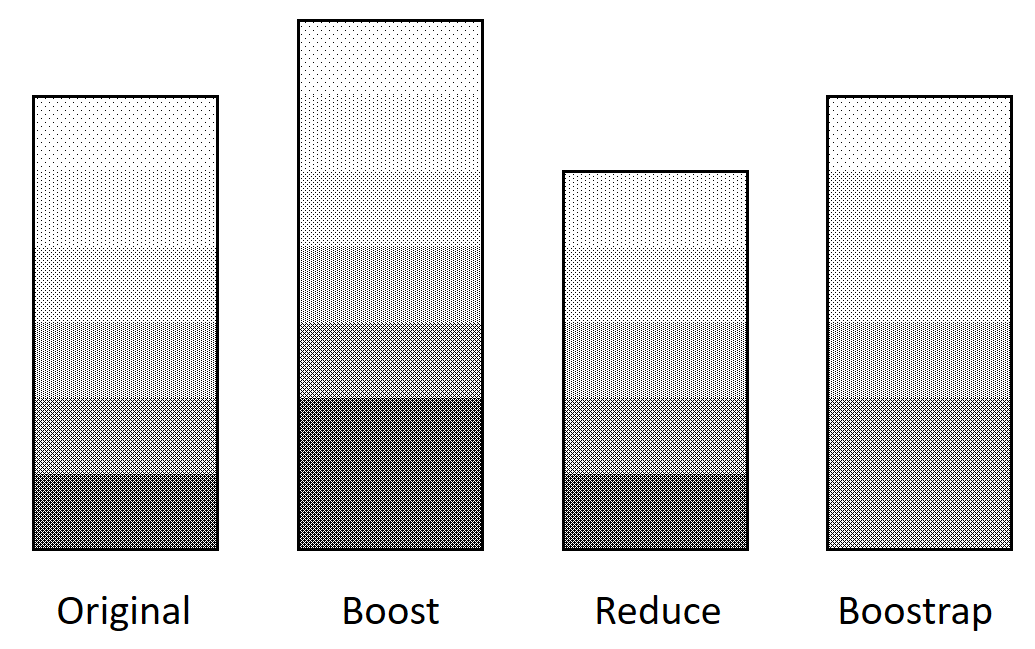}
\caption{Training data selection. Levels of grey are used to indicate perplexity ranges. Darker/lighter indicate higher/lower perplexity.}
\label{dataselection}
\end{figure}

Inspired by machine learning algorithms (e.g. boosting, bootstrap, etc.),
which are widely used to improve the stability and accuracy especially in the field of text classification,
we force the neural network to pay closer attention to ``difficult'' training examples.
To our knowledge, this is the first approach that integrates such meta-algorithms within NMT training.
Figure \ref{dataselection} illustrates different methods to select training instances. Instances are initially sorted based on their translation perplexities (Original) to be finally duplicated (Boost), reduced (Reduce), or just re-sampled (Bootstrap).
Such adjustment is applied during each training epoch, thus results in different cost and accuracy.

To be specific, at each training epoch we extend the training set with sentence pairs that the translation model finds ``difficult'' to translate ({\bf Boost}).
We approximate this procedure by choosing sentences with a high perplexity score. In Figure \ref{dataselection}, the block with higher perplexity sentences (darker) is repeated in the Boosting set when compared to the Original set.
The process has no additional computational cost since perplexity is already computed in a previous iteration.

To put it further, we may focus on ``difficult'' sentences by removing ``easy'' ones from the training set ({\bf Reduce}). In Figure \ref{dataselection}, the block with lower perplexity sentences (lighter) is missing in the Reduction set when compared to the Original set.

Finally, we randomly sample 100\% of the sentences from the corpus as a comparison to the baseline system ({\bf Bootstrap}). In Figure \ref{dataselection}, some blocks of the Original set appear repeated in the Bootstrap set while some others are missing, due to a random re-sampling.

\section{Experiments}

In this section we report on the experiments conducted to evaluate the suitability of the proposed methods. We begin with details of the NMT system parameters as well as the corpora employed.

\subsection{System Description}
\label{ssec:desc}
We build our NMT system based on an open-source project \texttt{OpenNMT}\footnote{\url{http://opennmt.net}}.
We use a bidirectional RNN encoder with $4$ LSTM layers with each containing $1,000$ nodes.
Word embeddings are sized of $500$ cells and we set the dropout probability to $0.3$.
Batch size is set to $64$.
The maximum length of both source and target sentences is set to $80$ and we limit the vocabulary size to $50K$ words for both source and target languages.

The default optimiser is SGD with starting learning rate $1.0$.
We start to decay the learning rate from epoch 10, or when we find a perplexity increasing compared with the previous epoch on a validation set.
Evaluation is performed on a held-out testset with BLEU score \cite{papineni2002bleu}. 

\subsection{Corpora}

We used an in-house generic corpus built as mix from several client data consisting of French-English sentences pairs.
We split the corpus into three sets: training, validation and a held-out test set.
Table \ref{tab:XP_DA} shows statistics of the data used in our experiments.

\begin{table} [h]
\centerline{
\scalebox{1}{
\begin{tabular}{|l|ccc|}
\hline
 & \#sents & \#tok (EN) & \#tok (FR) \\
\hline
Train & 1M     & 24M & 26M  \\
Valid & 2,000  & 48K & 55K  \\
Test  & 2,000 & 48K & 54K    \\
\hline
\end{tabular}}}
\caption{\label{tab:XP_DA} Statistics of the data used in our experiments. M stands for millions and K for thousands.}
\end{table}

\subsection{Results}
\label{ssec:results}

We train four systems corresponding to the different training policies considered in this paper following the system configuration detailed in Section \ref{ssec:desc}.
During each training epoch:

\begin{itemize}
\item {\bf default} uses the entire original data.
\item {\bf boost} extends 10\% the original data with the most difficult sentence pairs (following perplexity).
\item {\bf reduce} keeps 80\% the most difficult instances of the previous epochs, discarding the remaining 20\%. Note that the procedure restarts using the entire training set every 3 epochs. That is it uses 100\%, 80\%, 64\%, 100\%, 80\%, 64\%, ... of the training data.
\item {\bf bootstrap} re-samples 100\% the training set. Hence allowing for repeated and missing sentences of the original training set. 
\end{itemize}

All systems are trained up to 18 epochs\footnote{For some experiments, we continue to train until 22 epochs. However, there is no further improvement.}.
Evaluation results are shown in Figure \ref{fig:boosting_bleu} and Table \ref{tab:bleu}.
Each result is averaged from two systems initialised with different random seeds to alleviate the influence of randomisation.

As it can be seen, {\bf boost} outperforms the default method by $+1.49$ (BLEU) at the cost of using $10\%$ of additional training data. 
However, the system converges faster than any other system as best performance is achieved at epoch $14$,
while others need to achieve the best after epoch $16$.

\input{1M_single_shuffle.tex}


The {\bf reduce} system finally obtains the highest accuracy scores, outperforming the default method by $+1.63$ (BLEU).
In addition to accuracy, the system performing the {\bf reduce} method needed only $80\%$ less data than the default.
It shows that this policy is promising.



Considering {\bf boostrap}, the score steadily improves.
Though not so significantly, it outperforms also the default method by $+0.87$ (BLEU) with better stability.
Finally, we ensemble the 4 best systems (epoch 18) generated by each method, getting an additional $+0.92$ BLEU \mbox{improvement.}

\begin{table} [h]
\centerline{
\scalebox{1}{
\begin{tabular}{|l|cc|}
\hline
 & BLEU & Data \\
\hline
default & 52.49 & 100\% \\
boost & 53.98 & 110\% \\
reduce & {\bf 54.12} & {\bf 80\%} \\
boostrap & 53.36 & 100\% \\
\hline
Ensemble & 55.04 & - \\
\hline
\end{tabular}}}
\caption{\label{tab:bleu} BLEU score, and data size conditions for different training policies.}
\end{table}




\subsubsection{Perplexity Normalisation}

In this work we use perplexity as the measure for translation ``difficulty'', computed over each training batch.
We distinguish instances between ``difficult'' ones and ``easy'' ones in order to save time during system training.
Since perplexity will always increase when the sentences are long, a normalisation method is typically needed.
We test several normalisation strategies:
\begin{itemize}
\item by batch size (number of training instances)
\item by (target) sentence length
\item without normalisation (longer sentences were always assigned a higher perplexity)
\end{itemize}

Experimental results show no significant performance differences for any of the strategies. An explanation for such behaviour may be found on the fact that long sentences are also difficult translations. Thus, the selected subsets (by any normalisation strategy) are very similar.

\subsection{Analysis}

We propose a simple data manipulation technique to help improving efficiency and performance of NMT models during training.
The idea imitates the human learning process.
Typically, humans need to spend more time to learn ``difficult'' concepts and less time for easier ones .
As a consequence, we force the NMT system to spend more time on more ``difficult'' instances, while ``skipping'' easier examples at the same time.
Thus, We emulate a human spending additional energy on learning complex concepts.

We inspect the selected ``difficult'' examples according to perplexity.
We find that almost all such examples containing complex structures, thus being difficult to be translated.
To force the system to pay much attention on them can adjust it towards ``mastering'' more information for these sentences.



An interesting conclusion is that, we can train the NMT system with 80\% of the most complex sentences. 
That is to say, when training examples with smaller perplexity are removed and those with larger ones are emphasised, the system performs better in terms of accuracy and efficiency.

Further experiments need to be conducted for a detailed insight of the methods presented. Like measuring the impact of using several ratios of training data boosted/reduced. As well as studying the impact of the methods on different language pairs and data size conditions.

\section{Conclusions}

For NMT, training cost is a big problem for even a medium-sized corpus with cutting-edge hardware.
At the same time, the trained model is apt to converge to a local optima, which makes the training more instable.
In this paper, we proposed a data boosting method for NMT to help improving stability and efficiency.
Experiments show that the improvement is quite stable during almost all training iterations.
By adding 10\% training corpus, translation score is improved by 1.49 BLEU scores and by reducing the size of the corpus by 20\%, translation performance improved by 1.63.

The method we proposed focuses on training process only.
There is no restriction for the neural network structure.
It can be used in any data parallelism framework and then distributed onto multi-GPUs.
Also, corpus pre-processing like tokenization (e.g. using sub-word unit \cite{sennrich2015neural}) and other techniques like guided training \cite{shen-EtAl:2016:P16-1,chen2016guided} can be freely added based on the method we proposed.
In the future, we plan to investigate more on the influence of training data especially in the later phase of training.

\section*{Acknowledgments}

We thank the anonymous reviewers for their insightful comments.

\bibliography{ijcnlp2017}
\bibliographystyle{ijcnlp2017}

\end{document}

%% file: 1M_single_shuffle.tex
\begin{figure}
  \begin{tikzpicture}
    \begin{axis}[
      xlabel={Epoch},
      ylabel={BLEU},
      xmin=1, xmax=18,
      ymin=36, ymax=55,
      xtick={2,4,6,8,10,12,14,16,18},
      ytick={36,38,40,42,44,46,48,50,52,54},
      width=0.45\textwidth,
      ymajorgrids=true,
      grid style=dashed,
      legend pos=south east,
      legend cell align=left,
      legend style={font=\tiny, draw=none},
      ]
      \addplot[
      color=black,
      mark=triangle,
      ]
      coordinates {
(1, 38.85)
(2, 44.45)
(3, 45.58)
(4, 47.37)
(5, 48.15)
(6, 48.52)
(7, 49.15)
(8, 49.67)
(9, 49.36)
(10, 50.55)
(11, 50.83)
(12, 51.77)
(13, 52.31)
(14, 52.27)
(15, 52.44)
(16, 52.61)
(17, 52.59)
(18, 52.49)
      };
      \addplot[
      color=blue,
      mark=diamond,
      ]
      coordinates {
(1, 35.86)
(2, 45.42)
(3, 48.64)
(4, 48.83)
(5, 50.23)
(6, 50.96)
(7, 51.37)
(8, 52.56)
(9, 52.69)
(10, 53.37)
(11, 53.61)
(12, 53.77)
(13, 54.04)
(14, 54.09)
(15, 53.92)
(16, 53.95)
(17, 53.98)
(18, 53.98)
      };
      \addplot[
      color=green,
      mark=star,
      ]
      coordinates {
(1, 35.86)
(2, 43.84)
(3, 46.65)
(4, 47.3)
(5, 49.15)
(6, 50.25)
(7, 49.78)
(8, 50.7)
(9, 51.08)
(10, 51.66)
(11, 52.6)
(12, 53.18)
(13, 53.42)
(14, 53.96)
(15, 53.88)
(16, 54.15)
(17, 54.17)
(18, 54.12)
      };
      \addplot[
      color=red,
      mark=square,
      ]
      coordinates {
(1, 38.91)
(2, 43.94)
(3, 46.19)
(4, 46.47)
(5, 48.69)
(6, 48.95)
(7, 49.45)
(8, 50.31)
(9, 50.8)
(10, 51.04)
(11, 51.86)
(12, 52.11)
(13, 52.86)
(14, 52.73)
(15, 53.37)
(16, 53.45)
(17, 53.44)
(18, 53.36)
      };
      \legend{default, boost 10\%, reduce 20\% restart every 3 epochs, bootstrap 100\% for each epoch}
    \end{axis}
  \end{tikzpicture}
  \caption{Effect of data boosting during training for English-French translation.}
  \label{fig:boosting_bleu}
\end{figure}
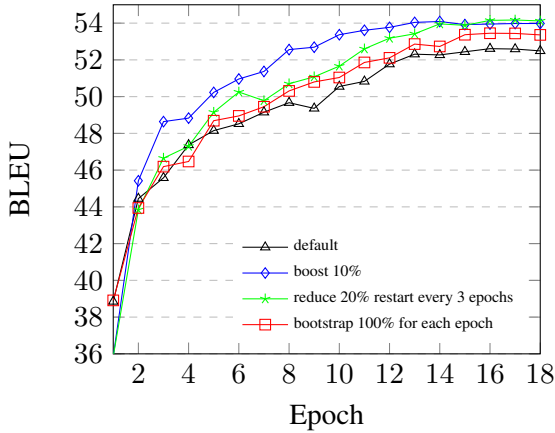

%% file: ijcnlp2017.bbl
\begin{thebibliography}{21}
\expandafter\ifx\csname natexlab\endcsname\relax\def\natexlab#1{#1}\fi

\bibitem[{Bahdanau et~al.(2014)Bahdanau, Cho, and Bengio}]{bahdanau2014neural}
Dzmitry Bahdanau, Kyunghyun Cho, and Yoshua Bengio. 2014.
\newblock Neural machine translation by jointly learning to align and
  translate.
\newblock \emph{arXiv preprint arXiv:1409.0473}.

\bibitem[{Chen et~al.(2016{\natexlab{a}})Chen, Matusov, Khadivi, and
  Peter}]{chen2016guided}
Wenhu Chen, Evgeny Matusov, Shahram Khadivi, and Jan-Thorsten Peter.
  2016{\natexlab{a}}.
\newblock Guided alignment training for topic-aware neural machine translation.
\newblock \emph{arXiv preprint arXiv:1607.01628}.

\bibitem[{Chen et~al.(2016{\natexlab{b}})Chen, Liu, Wang, Gales, and
  Woodland}]{chen2016efficient}
Xie Chen, Xunying Liu, Yongqiang Wang, Mark~JF Gales, and Philip~C Woodland.
  2016{\natexlab{b}}.
\newblock Efficient training and evaluation of recurrent neural network
  language models for automatic speech recognition.
\newblock \emph{IEEE/ACM Transactions on Audio, Speech, and Language
  Processing}, 24(11):2146--2157.

\bibitem[{Cho et~al.(2015)Cho, Memisevic, and Bengio}]{cho2015using}
S{\'e}bastien Jean~Kyunghyun Cho, Roland Memisevic, and Yoshua Bengio. 2015.
\newblock On using very large target vocabulary for neural machine translation.

\bibitem[{Crego et~al.(2016)Crego, Kim, Klein, Rebollo, Yang, Senellart,
  Akhanov, Brunelle, Coquard, Deng, Enoue, Geiss, Johanson, Khalsa, Khiari, Ko,
  Kobus, Lorieux, Martins, Nguyen, Priori, Riccardi, Segal, Servan, Tiquet,
  Wang, Yang, Zhang, Zhou, and Zoldan}]{DBLP:journals/corr/CregoKKRYSABCDE16}
Josep~Maria Crego, Jungi Kim, Guillaume Klein, Anabel Rebollo, Kathy Yang, Jean
  Senellart, Egor Akhanov, Patrice Brunelle, Aurelien Coquard, Yongchao Deng,
  Satoshi Enoue, Chiyo Geiss, Joshua Johanson, Ardas Khalsa, Raoum Khiari,
  Byeongil Ko, Catherine Kobus, Jean Lorieux, Leidiana Martins, Dang{-}Chuan
  Nguyen, Alexandra Priori, Thomas Riccardi, Natalia Segal, Christophe Servan,
  Cyril Tiquet, Bo~Wang, Jin Yang, Dakun Zhang, Jing Zhou, and Peter Zoldan.
  2016.
\newblock \href {http://arxiv.org/abs/1610.05540} {Systran's pure neural
  machine translation systems}.
\newblock \emph{CoRR}, abs/1610.05540.

\bibitem[{Gehring et~al.(2017)Gehring, Auli, Grangier, Yarats, and
  Dauphin}]{gehring2017convolutional}
Jonas Gehring, Michael Auli, David Grangier, Denis Yarats, and Yann~N Dauphin.
  2017.
\newblock Convolutional sequence to sequence learning.
\newblock \emph{arXiv preprint arXiv:1705.03122}.

\bibitem[{Hubara et~al.(2016)Hubara, Courbariaux, Soudry, El-Yaniv, and
  Bengio}]{hubara2016quantized}
Itay Hubara, Matthieu Courbariaux, Daniel Soudry, Ran El-Yaniv, and Yoshua
  Bengio. 2016.
\newblock Quantized neural networks: Training neural networks with low
  precision weights and activations.
\newblock \emph{arXiv preprint arXiv:1609.07061}.

\bibitem[{Ioffe and Szegedy(2015)}]{pmlr-v37-ioffe15}
Sergey Ioffe and Christian Szegedy. 2015.
\newblock Batch normalization: Accelerating deep network training by reducing
  internal covariate shift.
\newblock In \emph{Proceedings of the 32nd International Conference on Machine
  Learning}, volume~37 of \emph{Proceedings of Machine Learning Research},
  pages 448--456, Lille, France. PMLR.

\bibitem[{Jiang and Zhai(2007)}]{jiang2007instance}
Jing Jiang and ChengXiang Zhai. 2007.
\newblock Instance weighting for domain adaptation in nlp.
\newblock In \emph{ACL}, volume~7, pages 264--271.

\bibitem[{Kalchbrenner et~al.(2016)Kalchbrenner, Espeholt, Simonyan, Oord,
  Graves, and Kavukcuoglu}]{kalchbrenner2016neural}
Nal Kalchbrenner, Lasse Espeholt, Karen Simonyan, Aaron van~den Oord, Alex
  Graves, and Koray Kavukcuoglu. 2016.
\newblock Neural machine translation in linear time.
\newblock \emph{arXiv preprint arXiv:1610.10099}.

\bibitem[{Kim and Rush(2016)}]{kim2016sequence}
Yoon Kim and Alexander~M Rush. 2016.
\newblock Sequence-level knowledge distillation.
\newblock \emph{arXiv preprint arXiv:1606.07947}.

\bibitem[{Koehn and Knowles(2017)}]{koehn2017six}
Philipp Koehn and Rebecca Knowles. 2017.
\newblock Six challenges for neural machine translation.
\newblock \emph{arXiv preprint arXiv:1706.03872}.

\bibitem[{Kuchaiev and Ginsburg(2017)}]{kuchaiev2017factorization}
Oleksii Kuchaiev and Boris Ginsburg. 2017.
\newblock Factorization tricks for lstm networks.
\newblock \emph{arXiv preprint arXiv:1703.10722}.

\bibitem[{Papineni et~al.(2002)Papineni, Roukos, Ward, and
  Zhu}]{papineni2002bleu}
Kishore Papineni, Salim Roukos, Todd Ward, and Wei-Jing Zhu. 2002.
\newblock Bleu: a method for automatic evaluation of machine translation.
\newblock In \emph{Proceedings of the 40th annual meeting on association for
  computational linguistics}, pages 311--318. Association for Computational
  Linguistics.

\bibitem[{See et~al.(2016)See, Luong, and Manning}]{see2016compression}
Abigail See, Minh-Thang Luong, and Christopher~D Manning. 2016.
\newblock Compression of neural machine translation models via pruning.
\newblock \emph{arXiv preprint arXiv:1606.09274}.

\bibitem[{Sennrich et~al.(2015)Sennrich, Haddow, and
  Birch}]{sennrich2015neural}
Rico Sennrich, Barry Haddow, and Alexandra Birch. 2015.
\newblock Neural machine translation of rare words with subword units.
\newblock \emph{arXiv preprint arXiv:1508.07909}.

\bibitem[{Shen et~al.(2016)Shen, Cheng, He, He, Wu, Sun, and
  Liu}]{shen-EtAl:2016:P16-1}
Shiqi Shen, Yong Cheng, Zhongjun He, Wei He, Hua Wu, Maosong Sun, and Yang Liu.
  2016.
\newblock Minimum risk training for neural machine translation.
\newblock In \emph{Proceedings of the 54th Annual Meeting of the Association
  for Computational Linguistics (Volume 1: Long Papers)}, pages 1683--1692,
  Berlin, Germany. Association for Computational Linguistics.

\bibitem[{Sutskever et~al.(2014)Sutskever, Vinyals, and
  Le}]{sutskever2014sequence}
Ilya Sutskever, Oriol Vinyals, and Quoc~V Le. 2014.
\newblock Sequence to sequence learning with neural networks.
\newblock In \emph{Advances in neural information processing systems}, pages
  3104--3112.

\bibitem[{Vaswani et~al.(2017)Vaswani, Shazeer, Parmar, Uszkoreit, Jones,
  Gomez, Kaiser, and Polosukhin}]{1706.03762}
Ashish Vaswani, Noam Shazeer, Niki Parmar, Jakob Uszkoreit, Llion Jones,
  Aidan~N. Gomez, Lukasz Kaiser, and Illia Polosukhin. 2017.
\newblock \href {http://arxiv.org/abs/arXiv:1706.03762} {Attention is all you
  need}.

\bibitem[{Wu et~al.(2016)Wu, Schuster, Chen, Le, Norouzi, Macherey, Krikun,
  Cao, Gao, Macherey et~al.}]{wu2016google}
Yonghui Wu, Mike Schuster, Zhifeng Chen, Quoc~V Le, Mohammad Norouzi, Wolfgang
  Macherey, Maxim Krikun, Yuan Cao, Qin Gao, Klaus Macherey, et~al. 2016.
\newblock Google's neural machine translation system: Bridging the gap between
  human and machine translation.
\newblock \emph{arXiv preprint arXiv:1609.08144}.

\bibitem[{Zhou et~al.(2016)Zhou, Cao, Wang, Li, and Xu}]{zhou2016deep}
Jie Zhou, Ying Cao, Xuguang Wang, Peng Li, and Wei Xu. 2016.
\newblock Deep recurrent models with fast-forward connections for neural
  machine translatin.
\newblock \emph{arXiv preprint arXiv:1606.04199}.

\end{thebibliography}
